\documentclass[lettersize,journal]{IEEEtran}
\usepackage{amsmath,amsfonts}
\usepackage{array}
\usepackage[caption=false,font=normalsize,labelfont=sf,textfont=sf]{subfig}
\usepackage{textcomp}
\usepackage{stfloats}
\usepackage{url}
\usepackage{verbatim}
\usepackage{graphicx}
\usepackage{cite}
\usepackage[ruled]{algorithm2e}
\usepackage{amssymb}

\usepackage[pagebackref,breaklinks]{hyperref}

\usepackage[capitalize]{cleveref}
\crefname{section}{Sec.}{Secs.}
\Crefname{section}{Section}{Sections}
\Crefname{table}{Table}{Tables}
\crefname{table}{Tab.}{Tabs.}

\usepackage{booktabs}
\usepackage{pifont}
\usepackage{multirow}
\usepackage{bm}
\usepackage[table,xcdraw]{xcolor}

\usepackage{dsfont}
\usepackage{graphics}
\usepackage[export]{adjustbox}

\begin{document}

\title{SemiCD-VL: Visual-Language Model Guidance Makes Better Semi-supervised Change Detector}

\author{Kaiyu Li, Xiangyong Cao, Yupeng Deng, Jiayi Song, Junmin Liu, Deyu Meng, Zhi Wang
\thanks{This work is partially supported by the National Key R\&D Program of China (2021ZD0112902), and China NSFC projects under contract 62272375, 12226004.\textit{(Corresponding author: Xiangyong Cao)}}
\thanks{Kaiyu Li and Zhi Wang are with the School of Software Engineering, Xi’an Jiaotong University, Xi’an 710049, China (email: likyoo.ai@gmail.com, zhiwang@xjtu.edu.cn)}
\thanks{Xiangyong Cao and Jiayi Song are with the School of Computer Science and Technology and Ministry of Education Key Lab For Intelligent Networks and Network Security, Xi’an Jiaotong University, Xi’an 710049, China (email: caoxiangyong@xjtu.edu.cn, songyangyifei@gmail.com)}
\thanks{Yupeng Deng is with the Aerospace Information Research Institute, Chinese Academy of Sciences, Beijing 100094, China (email: dengyp@aircas.ac.cn)}
\thanks{Junmin Liu and Deyu Meng are with the School of Mathematics and Statistics and Ministry of Education Key Lab of Intelligent Networks and Network Security, Xi’an Jiaotong University, Xi’an, Shaanxi, China, and Pazhou Laboratory (Huangpu), Guangzhou, Guangdong, China. (email: junminliu@mail.xjtu.edu.cn, dymeng@mail.xjtu.edu.cn).}
}

\markboth{Journal of \LaTeX\ Class Files,~Vol.~14, No.~8, August~2023}%
{Shell \MakeLowercase{\textit{et al.}}: Bare Demo of IEEEtran.cls for IEEE Journals}



\maketitle

\renewcommand{\thefootnote}{\fnsymbol{footnote}}

\begin{abstract}
Change Detection (CD) aims to identify pixels with semantic changes between images. However, annotating massive numbers of pixel-level images is labor-intensive and costly, especially for multi-temporal images, which require pixel-wise comparisons by human experts. Considering the excellent performance of visual language models (VLMs) for zero-shot, open-vocabulary, etc. with prompt-based reasoning, it is promising to utilize VLMs to make better CD under limited labeled data. In this paper, we propose a VLM guidance-based semi-supervised CD method, namely SemiCD-VL. The insight of SemiCD-VL is to synthesize free change labels using VLMs to provide additional supervision signals for unlabeled data. However, almost all current VLMs are designed for single-temporal images and cannot be directly applied to bi- or multi-temporal images. Motivated by this, we first propose a VLM-based mixed change event generation (CEG) strategy to yield pseudo labels for unlabeled CD data. Since the additional supervised signals provided by these VLM-driven pseudo labels may conflict with the original pseudo labels from the consistency regularization paradigm (e.g. FixMatch), we propose the dual projection head for de-entangling different signal sources. Further, we explicitly decouple the bi-temporal images semantic representation through two auxiliary segmentation decoders, which are also guided by VLM. Finally, to make the model more adequately capture change representations, we introduce contrastive consistency regularization by constructing feature-level contrastive loss in auxiliary branches. Extensive experiments show the advantage of SemiCD-VL. For instance, SemiCD-VL improves the FixMatch baseline by +5.3 $IoU^c$ on WHU-CD and by +2.4 $IoU^c$ on LEVIR-CD with 5\% labels, and SemiCD-VL requires only 5\% to 10\% of the labels to achieve performance similar to the supervised methods. In addition, our CEG strategy, in an un-supervised manner, can achieve performance far superior to state-of-the-art (SOTA) un-supervised CD methods (e.g., IoU improved from 18.8\% to 46.3\% on LEVIR-CD dataset). Code is available at \url{https://github.com/likyoo/SemiCD-VL}. 
\end{abstract}

\begin{IEEEkeywords}
Change detection, Semi-supervised learning, Vision-language model, Foundation model.
\end{IEEEkeywords}

\section{Introduction}\label{sec:1}

\par Change detection (CD) is a fundamental task in the practice of Earth observation \cite{fang2023changer,li2024new,daudt2018fully}, industrial quality control \cite{park2021changesim,park2022dual}, autonomous driving \cite{alcantarilla2018street,lei2020hierarchical}, robotics \cite{salimpour2022self}, etc., aiming at identifying changes at the pixel level between images. However, pixel-level annotation is labor-intensive and costly, especially for these CD tasks, which require human experts to carefully compare pixel-level changes between image pairs, making the annotation more difficult \cite{zheng2021change, zhang2024boosting}. Hence there is an urgent need for semi-supervised or un-supervised methods to mitigate the reliance on labeled data for CD tasks.

\par Compared with supervised CD, semi-supervised and un-supervised CD only needs little or no labeled data for training, which is closer to real scenarios and thus has higher practical applications \cite{yang2022survey}. Especially for semi-supervised CD, as a trade-off between supervised and un-supervised CD, can potentially achieve close performance to supervised CD with an acceptable annotation volume ($\sim$ 10\% of supervised manner) \cite{yang2023revisiting}. Based on some assumptions (e.g., generative model assumption \cite{goodfellow2014generative}, low density assumption \cite{learning2006semi}, etc.), semi-supervised CD attempts to build reasonable supervised signals (e.g. pseudo label) for a large amount of free unlabeled data and thus enhance the feature representation of the model \cite{yang2022survey}. 
Additionally, for the semi-supervised CD task, two critical factors need to be considered for building pseudo labels of unlabeled data \cite{wang2022semi}: 1) the reliability of the pseudo labels, where unreliable pseudo labels may directly lead to misguidance and accumulation of errors; 2) the abundance and diversity of the pseudo labels, where too sparse supervised signals bring limited gain since the CD task requires dense supervision. Both of these aspects have been explored to some extent in the general semi-supervised progression. For instance, previous studies acquire more reliable pseudo labels through adversarial method \cite{hung2018adversarial, mittal2019semi, ke2020guided}, contrastive learning \cite{wang2022semi}, threshold control, etc., and more diverse supervision through multiple teacher networks \cite{na2024switching}, etc. In this paper, we propose to use a vision-language model (VLM) to generate extra reliable pseudo labels to facilitate the semi-supervised CD, which is a new attempt for the CD task in the VLM era and takes into account both of the above factors.

\par VLM refers to the model that can process and understand the modalities of language (text) and vision (image) simultaneously to perform advanced vision-language tasks, e.g. vision grounding, image captioning, etc. In recent years, numerous VLMs have emerged that revolutionize the conventional pre-training, fine-tuning, and prediction paradigms, allowing for zero-shot or open-vocabulary (OV) recognition, i.e., performing a wide range of tasks with textual prompts. However, almost all current VLMs are designed for single-temporal images, for instance, in CLIP-based models, ``a photo of a \{object\}.'' can be used as a prompt to localize the object of interest. However, it is difficult to directly define a vocabulary or sentence that represents a change, e.g. ``difference'', as a prompt for VLM, because such a definition is abstract. Motivated by this, we first propose a VLM-based mixed change event generation (CEG) strategy, which avoids abstract definitions and generates pseudo change labels in an OV fashion. 

\par
\begin{figure}[t]
  \centering
   \includegraphics[width=1.0\linewidth]{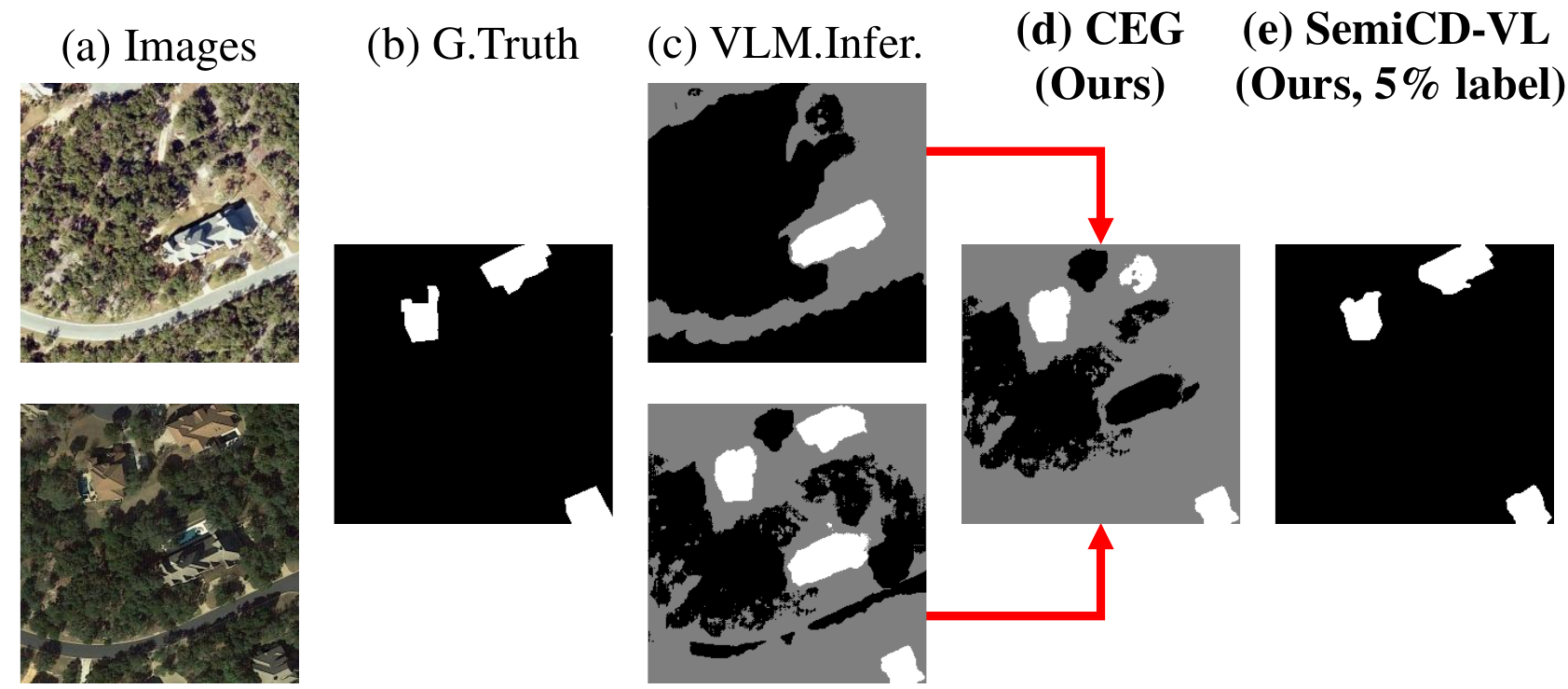}
   \caption{Inputs and outputs of SemiCD-VL. (a) and (b) represent the bi-temporal images and their change label. (c) denotes the semantic segmentation masks from VLM's single-temporal image reasoning, and we use the mixed CEG algorithm to convert them into change mask (d) as the supplementary supervised signal. (e) is the prediction of SemiCD-VL after semi-supervised training (white rendering indicates pixels with semantic changes, black indicates no semantic changes, and gray indicates unreliable regions, which are ignored in the loss computation).}
   \label{fig:fig1}
\end{figure}

 The overall architecture of our method is based on consistency regularization in semi-supervised training, i.e., FixMatch \cite{sohn2020fixmatch}. With the inclusion of the VLM guidance, an obvious issue is the potential conflict between the supervised signals from the VLM and the original supervised signals from the consistency regularization for strong perturbation predictions. For this, we design a dual projection head to de-entangle the different signal sources. During the generation of change labels by mixed CEG, semantic masks for single-temporal images are also generated, as shown in Fig. \ref{fig:fig1}(c), and they can be employed to encourage the model to perform semantic segmentation of single-temporal images, which explicitly decouples the process of CD. Additionally, this allows metric-aware supervision based on feature-level contrastive loss to be part of the consistency regularization.


To the best of our knowledge, our method, called SemiCD-VL, is the first work to utilize the VLM guidance for semi-supervised CD. Specifically, our SemiCD-VL contains five components:

\begin{itemize}

\item Mixed CEG: To build more diverse and reliable supervised signals, we propose mixed CEG which combines pixel-level CEG and instance-level CEG. Mixed CEG can avoid misalignment error and filter out low-confidence predictions of VLM.

\item VLM guidance: We build uniform VLM supervisions for unlabeled samples with different degrees of perturbation, which provides more diverse supervised signals and implicitly maximizes the similarity between different views of the same sample.

\item Dual projection head: To avoid potential conflicts that exist in the pseudo labels generated by the consistency regularization paradigm and VLM, we propose the dual projection head to de-entangle different supervised signal sources.

\item Decoupled semantic guidance: In addition to the change masks, the VLM can infer the individual semantic segmentation mask for each temporal image, which provides an additional supervised signal for CD and an explicit way to decouple the semantic representations of the bi-temporal images. Specifically, we add two auxiliary semantic decoders to the model, which generate their respective semantic masks.

\item Contrastive consistency regularization: To make the model capture change representations more efficiently, we introduce metric-aware supervision via feature-level contrastive loss in two auxiliary decoders. The feature-level contrastive loss aims to pull the distance between feature vectors with semantic similarity closer together and push those with semantic changes farther apart.

\end{itemize} 

All five components are applied only at training time and do not introduce additional inference overhead. We evaluated SemiCD-VL on two large public remote sensing change detection (RSCD) datasets. The experimental results confirm the superiority of the proposed method over other semi-supervised methods. The detailed ablation studies suggest that the proposed components are significant. Surprisingly, the proposed CEG strategy, achieves state-of-the-art (SOTA) un-supervised CD performance and far superior to other methods.

The rest of this paper is organized as follows. Section \ref{sec:rel_work} briefly introduces supervised CD, semi-supervised learning and dense prediction with VLMs. Section \ref{sec:method} presents the proposed semi-supervised framework and the implementation details. The extensive experimental results and detailed discussion are presented in Section \ref{sec:exp}. Finally, our conclusions are given in Section \ref{sec:concl}.

\section{Related Works}
\label{sec:rel_work}

\subsection{Supervised Change Detection}

\par Different from scene interpretation of single-temporal images, CD needs to additionally model event changes in bi-temporal images, which is involved in a wide range of real-world scenarios. To achieve more robust system navigation and planning in self-driving, Alcantarilla et al. \cite{alcantarilla2018street} proposed the street-view change detection model under monocular camera for efficient map maintenance. Also in the smart city environment, Varghese et al. \cite{varghese2018changenet} proposed ChangeNet, which utilizes drone photography to achieve automatic monitoring of changes in the urban context and improve the management of public infrastructure.

\par In industrial scenarios, by comparing the current state with the previous state or the standard state, CD can identify device failures, device movements, product defects, etc. Taking into account both pairing and detection, Park et al. \cite{park2021changesim} proposed ChangeSim, a dataset for the detection of changes in the online scene in indoor industrial environments. Subsequently, SimSaC \cite{park2022dual} was proposed, which concurrently conducts scene flow estimation and change detection to alleviate the problem of imperfect matching of bi-temporal images.

\par The most active field in which CD is applied is remote sensing (RS) imagery, based on the fact that in long-term earth observation, it is common to focus only on the categories of land cover in the changed area, rather than repeatedly observing all pixels in the whole area \cite{fang2023changer}. In recent years, many excellent works have emerged for the RSCD task; these methods use siamese encoders to extract bi-temporal features and use a binary segmentation head to compute change/unchanged probabilities \cite{daudt2018fully, chen2020spatial, fang2021snunet, fang2023changer}. In addition, some methods use temporal-wise semantic segmentation as the auxiliary task, aiming to decouple the change process and establish more explicit supervision signals. Typically, based on the inductive bias of the causal relationship of semantic changes and temporal symmetry, Zheng et al. \cite{zheng2022changemask} proposed a general encoder transformer-decoder framework, ChangeMask, for the detection of semantic changes and specifically introduced a temporal symmetric transformer to interact and fuse bitemporal features. Tian et al. \cite{tian2023temporal} used sensitive objects in single-temporal images to enhance the spatio-temporal features, and the proposed TCRPN can plug-and-play to other models e.g., ChangeMask, and segment more refined regions of change.

\par The above-mentioned methods are trained on sufficient labeled data and evaluated on the corresponding testing data. Although they have achieved some success, there is a lack of exploration under limited labeled data, i.e., few labels or even no labels, which is the focus of this paper.

\subsection{Semi-supervised Leanring}

\par Considering that pixel-level annotation is costly, semi-supervised learning (SSL) allows us to build models using both labeled and unlabeled data. Compared to pure supervised learning, SSL can enhance the representation capability of model through unlabeled data. Depending on the learning policy of unlabeled data, SSL can be divided into adversarial method \cite{souly2017semi, hung2018adversarial, mittal2019semi, ke2020guided}, pseudo labeling method \cite{yang2022st++}, consistency regularization method \cite{ouali2020semi} and their hybrid methods \cite{sohn2020fixmatch, yang2023revisiting, yang2022survey}.

\par The critical challenge in SSL is how to make full use of a large amount of unlabeled data and build reliable and abundant supervision signals to improve the generalizability of the model. To build reliable supervision signals, Hung et al. \cite{hung2018adversarial} and Mittal et al. \cite{mittal2019semi} introduced an extra discriminator network to generate a confidence map and constructed the learning process for unlabeled data by binarizing the reliable regions in the confidence map with a defined threshold. Correspondingly, Ke et al. \cite{ke2020guided} used a flaw detector to suppress unreliable prediction regions. In addition, to further utilize these unreliable prediction regions, i.e., low-quality pseudo labels, Wang et al. \cite{wang2022semi} introduced the contrastive learning method with unreliable pseudo labels as negative samples. Using the patch-wise CutMix strategy, Fang et al. \cite{fang2023locating} replaces unreliable regions with high entropy in unlabeled images with determined regions in labeled images. In general, these methods establish more reliable supervision for unlabeled data from different perspectives. On the other hand, some work has begun to focus on the diversity of supervised signals. Na et al. \cite{na2024switching} proposed Dual Teacher, which uses temporary teachers to periodically take turns generating pseudo labels to train a student model, and ensures teacher diversity through different perturbations and periodic exponential moving average (EMA). Based on FixMatch \cite{sohn2020fixmatch} and UniMatch \cite{yang2023revisiting}, Hoyer et al. \cite{hoyer2023semivl} introduced VLM to semi-supervised segmentation for the first time and proposed SemiVL to alleviate the confusion of classes with similar visual appearance under limited supervision. However, for CD tasks, SemiVL still has some issues: (1) Almost all current VLMs are designed for single-temporal images and cannot be directly applied to bi- or multi-temporal images. (2) Supervision from consistency regularization (weak perturbation) and supervision from VLM may be inconsistent, leading to confusing learning. (3) Since it is trained at the image level, CLIP \cite{radford2021learning} is sub-optimal in handling fine-grained tasks.

\par In this study, we introduce VLM for the semi-supervised CD task for the first time and address the above issues by our SemiCD-VL (corresponding to some components, i.e., mixed CEG strategy, dual projection head, and fine-grained VLM). Further, some strategies proposed in this paper are not limited to the CD task, but can also be applied to general semi-supervised methods.

\subsection{Dense Prediction with VLMs}

\par VLM connects knowledge from both image and text modalities and is a promising path toward general intelligence. Compared to the supervised/un-supervised pre-training, fine-tuning, and prediction paradigm of visual recognition, the new learning paradigm evoked by VLM enables efficient use of large-scale web data and zero-shot predictions that do not require task-specific fine-tuning. CLIP \cite{radford2021learning} demonstrates the power of visual language contrastive representation learning, which opens up new possibilities for visual recognition. Typically, OV detection/segmentation utilizes the VLM to extend knowledge, aiming to detect/segment targets of arbitrary textual descriptions (i.e., targets of any class beyond the base class). \cite{ding2022decoupling}, \cite{ghiasi2022scaling} and \cite{xu2022simple} proposed to solve the OV segmentation problem using two-stage methods, i.e., class-agnostic mask generation followed by mask classification using CLIP. MaskCLIP \cite{zhou2022extract} simplified this process in that it produces dense localized features by using CLIP's feature maps instead of global representation, and these features are then compared with the text embeddings to obtain fine-grained predictions. Additionally, ZegCLIP \cite{zhou2023zegclip} used both the class token and the dense features and generated more precise predictions with slight training.

On the other hand, Shen et al. \cite{APE} proposed a universal visual perception model, APE, which can perform detection, grounding, and segmentation tasks in one model. Specifically, APE uses a DETR-like \cite{carion2020end} model to generate region proposals and then uses an alignment head to select regions that correspond to vocabularies/sentences. APE is trained on several public datasets consisting mainly of natural images, and in this paper, we find that APE has strong generalization capabilities on optical RS images. Therefore, we use APE as the VLM in our pipeline by default.

\begin{figure*}[t]
  \centering
   \includegraphics[width=0.95\linewidth]{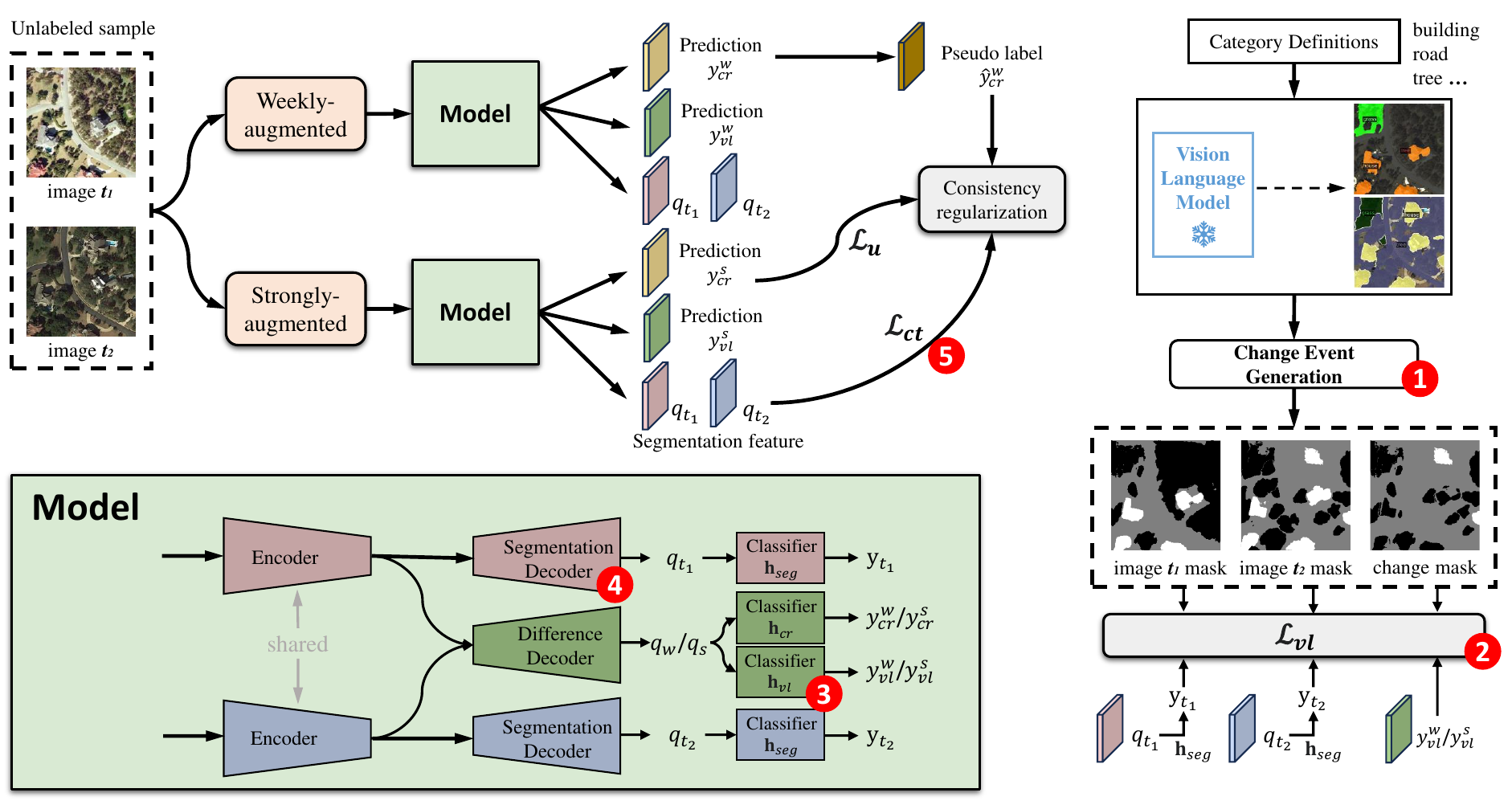}
   \caption{Overview of our \textbf{SemiCD-VL} framework. Utilizing the rich semantic representation of VLM, we propose 5 strategies (highlighted in red) to guide semi-supervised CD: We introduce the mixed CEG strategy in (1) that combines pixel-level CEG and instance-level CEG to generate reliable change masks, which guide the learning of unlabeled samples in (2). To avoid conflicts with supervised signals under the consistency regularization framework, dual projection heads are introduced in (3). Then, two auxiliary segmentation decoders are activated during the training phase to decouple the process of change prediction in (4), also benefiting from VLM guidance. Finally, contrastive consistency regularization is applied in (5) to make the model capture the change representation more explicitly. \includegraphics[scale=0.15,valign=c]{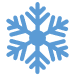} denotes the weights are frozen. For clarity, the features of the segmentation decoders for both weak and strong perturbations are denoted by $q_{t_1}$ and $q_{t_2}$. Components (1)-(5) correspond to Sections \ref{sec:mix_ceg} to \ref{sec:ccr}.}
   \label{fig:SemiCD-VL_detail}
\end{figure*}

\section{Method}
\label{sec:method}

\par In this section, we first briefly introduce the definition of semi-supervised CD and our baseline method FixMatch. Then we present the five components of SemiCD-VL in sub-section \ref{sec:mix_ceg} to \ref{sec:ccr}, as shown in Fig. \ref{fig:SemiCD-VL_detail}. Finally, we briefly present the construction details of our model. Indeed, in this research, we aim to demonstrate that VLM can facilitate CD learning under limited data, and don't suggest focusing too much on the model details.

\subsection{Preliminaries}
\label{sec:Preliminaries}

\par In Semi-supervised CD, we have a small set of labeled data $\mathcal{D}_{l}=\{\mathbf{x}_{i}^{l}, \mathbf{y}_{i}^{l}\}_{i=1}^{N_{l}}$ and a large set of unlabeled data $\mathcal{D}_{u}=\{\mathbf{x}_{i}^{u}\}_{i=1}^{N_{u}}$, where $\mathbf{x}_{i}^{*}=(\mathbf{t_1}_{i}^{*}, \mathbf{t_2}_{i}^{*})$ and $* \in \{l,u\}$, $\mathbf{t_1}_{i}^{*}$ and $\mathbf{t_2}_{i}^{*}$ denote the \textit{pre-event} and \textit{post-event} images, $N_{l}$ and $N_{u}$ denote the number of labeled and unlabeled images and $N_{u} \gg N_{l}$. Following the current mainstream weak-to-strong consistency regularization methods (e.g. FixMatch \cite{sohn2020fixmatch}) in SSL, we train our model on both labeled and unlabeled data. Specifically, the overall loss $\mathcal{L}_{cr}$ under consistency regularization framework consists of two cross-entropy (CE) loss terms, i.e., supervised loss $\mathcal{L}_{s}$ and un-supervised loss $\mathcal{L}_{u}$:

\begin{equation}
\begin{aligned}
  \mathcal{L}_{cr}=\frac{1}{2}(\mathcal{L}_{s}+\mathcal{L}_{u}).
  \label{eq:loss_overall}
\end{aligned}
\end{equation}
$\mathcal{L}_{s}$ is a standard pixel-wise CE loss, which is calculated on weakly perturbed labeled samples:

\begin{equation}
\begin{aligned}
  \mathcal{L}_{s}=\frac{1}{B_{l}} \sum \mathcal{H}(\mathbf{y}^{l}, f_\theta(\mathcal{A}^w(\mathbf{x}^{l}))),
  \label{eq:loss_s}
\end{aligned}
\end{equation}
where $B_{l}$ denotes the batch size of the labeled data, $f$ denotes the CD model, $\theta$ denotes its parameters, and $\mathcal{A}^w$ denotes weak perturbation. Based on the smoothing assumption \cite{4787647}, the consistency loss $\mathcal{L}_u$ is applied to encourage consistent predictions for the same image pair after different intensity perturbations:

\begin{equation}
\begin{aligned}
  \mathcal{L}_{u}=\frac{1}{B_{u}} \sum \mathds{1}(\max (\mathbf{y}^{w}) \geq \tau) \mathcal{H}(\hat{\mathbf{y}}^{w}, f_\theta(\mathcal{A}^s(\mathcal{A}^w(\mathbf{x}^{u})))),
  \label{eq:loss_u}
\end{aligned}
\end{equation}
where $B_{u}$ denotes the batch size of unlabeled data, 
$\mathcal{A}^s$ denotes strong perturbation, $\mathds{1}$ denotes the indicator function, $\tau$ denotes the confidence threshold, $\mathbf{y}^{w}$ denotes the prediction of weakly perturbed images, i.e. $\mathbf{y}^{w}=f_\theta(\mathcal{A}^w(\mathbf{x}^{u}))$, and $\hat{\mathbf{y}}^{w}$ denotes the hard label of $\mathbf{y}^{w}$, which can be obtained by $argmax()$ function.

\subsection{Change Event Generation}
\label{sec:mix_ceg}

\par To implement the semi-supervised CD under the guidance of VLM, the main problem faced is \textbf{how to use VLM to generate the pseudo change label}. We consider two possible schemes. The first is to use prompts directly to match the changed regions as in VLM reasoning for single temporal images \cite{zhou2022extract, zhou2023zegclip}. In this scheme, bi-temporal images are fed into the visual encoder and their visual features are fused, then the designed prompts are fed into the text encoder, and finally the textual embeddings and the fused local visual embeddings are matched. The advantage of this scheme is that it is an end-to-end process and does not require additional post-processing, however, the design of its prompts is difficult. In VLM, we generally use prompts representing objects as inputs to the text encoder, but the vocabularies representing changes are usually abstract, e.g., ``difference'', ``appear'', ``disappear'', etc., which makes it difficult to drive VLM to output the desired results. Another scheme is to decompose the bi-temporal change label generation into single-temporal images reasoning and conversion of their prediction masks into a change mask. We find that the latter one is more feasible and further propose mixed CEG under this scheme, which contains the following components, i.e., category definition, single-temporal image reasoning, pixel-level CEG and instance-level CEG. Next, we will introduce each component in detail.

\subsubsection{Category Definition \& Single-temporal Reasoning}

\begin{figure}[t]
  \centering
   \includegraphics[width=1.0\linewidth]{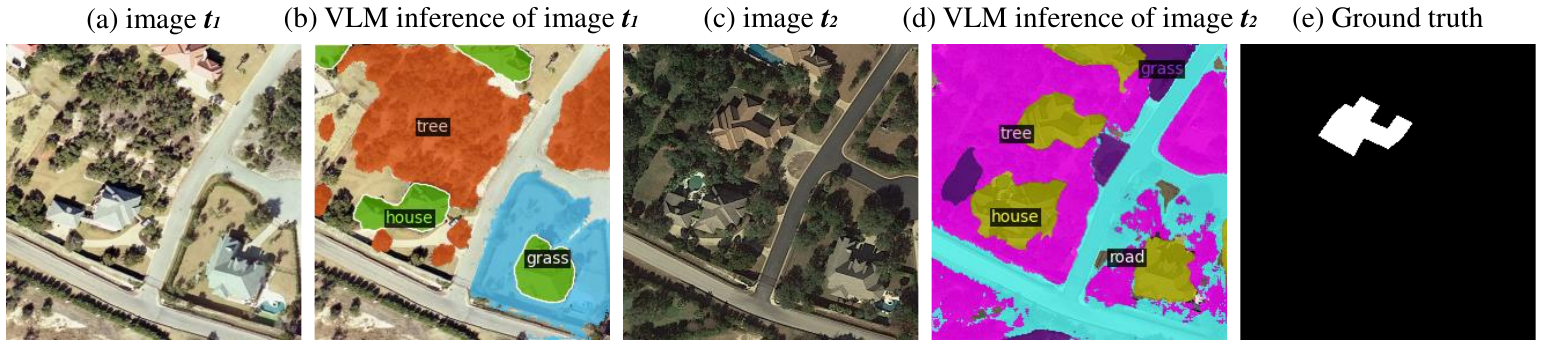}
   \caption{Visualization of direct inference using VLM with prompts: house, building, road, grass, tree, water. (The color rendering is random, just to distinguish different categories.) \protect\footnotemark[1]}
   \label{fig:bi_infer}
   \vspace{-1em}
\end{figure}

The definition of categories depends on the specific dataset, and in this paper, we evaluate our method on two public RSCD datasets, i.e., LEVIR-CD \cite{chen2020spatial} and WHU-CD \cite{ji2018fully}. We randomly select and observe some samples from the two datasets respectively, and find some of their consistent properties, e.g., they both focus mainly on the change of buildings. Therefore, we define the same category set for both datasets. Specifically, we define $\{\textit{house, building}\}$ for the \textbf{Foreground} and $\{\textit{road, grass, tree, water}\}$ for the \textbf{Background}, as shown in Fig. \ref{fig:bi_infer}.


\footnotetext[1]{powered by: \url{https://huggingface.co/spaces/shenyunhang/APE}}

\begin{figure}[t]
  \centering
   \includegraphics[width=1.0\linewidth]{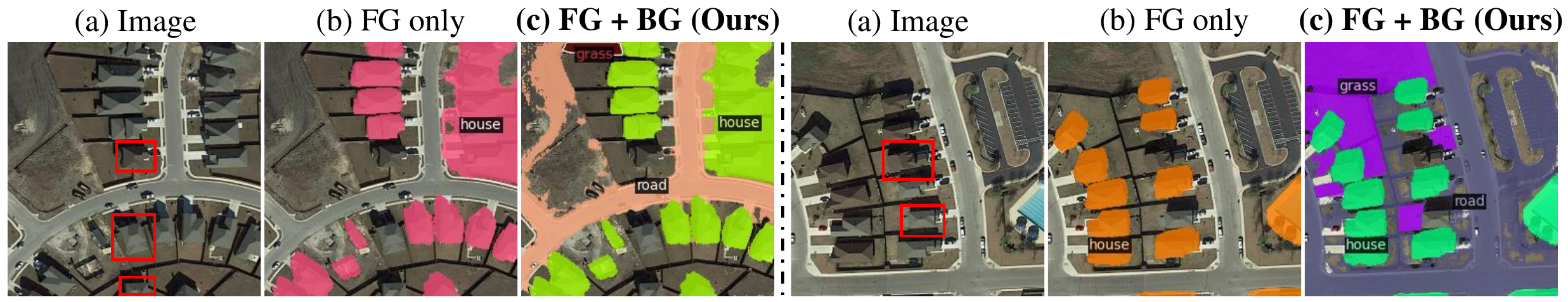}
   \caption{The influence of category definition on VLM reasoning. (b) denotes the prediction mask when only foreground categories are defined, and (c) denotes the prediction mask when categories for both foreground and background are defined. The red frames in (a) indicate targets that are incorrectly assigned to the background when only the foreground category is defined. (The color rendering is random, just to distinguish different categories.)}
   \label{fig:fig3}
   \vspace{-1em}
\end{figure}


\par It is worth noting that, different from general VLM-based applications, in addition to the definition of the foreground, we also explicitly define the category of the background. This is because RSCD is a black-or-white binary classification task (``changed'' or ``unchanged''), and the predictions of the two categories are mutually exclusive. However, in VLM's reasoning, there are uncertain foreground regions that may be assigned as background, leading to incorrect supervised signal. With the definition of the background category, both explicit foreground and background regions are acquired and the uncertain regions will be ignored. As shown in Fig. \ref{fig:fig3}, when using only foreground category for reasoning, some unrecognized buildings and houses are assigned as the background, and when there is a clear definition of the background category, these unrecognized regions are ignored and do not participate in the training process.

\par Next, each single-temporal image and the defined categories are fed into the VLM, e.g., ZegCLIP, APE, etc., to get the predicted probability map $P_{t}\in \mathbb{R}^{C \times H \times W}$, where $t \in \{1, 2\}$, $C$ denotes the sum of the number of foreground and background categories, and $H$ and $W$ denote the height and width of the original image. Here, we propose two strategies to generate change mask using $P_{t}$, i.e., pixel-level CEG and instance-level CEG.

\subsubsection{Pixel-level CEG}

\par As a basic CEG strategy, pixel-level CEG only considers the prediction $P_{t}(k) \in \mathbb{R}^{C \times 1 \times 1}$ for the position itself, where $P_{t}(k)$ denotes the probability vector of point $k$, and $k \in [1, H \times W]$. As mentioned above, we build a set of categories $B_a$ for each concept $a$ according to our category definitions, where $a \in A $ and $A = \{\textit{``Foreground''}, \textit{``Background''}\}$ here. For instance, the class ``house'' belongs to the concept ``Foreground'' (i.e. $house \in B_{Foreground}$). For each position $k$, we expect the concept probabilities $P^{\prime}_{t}(k,a)$. Here, the category with the highest score determines the predicted concept at position $k$, which can be formulated as:

\begin{equation}
\begin{aligned}
  P^{\prime}_{t}(k,a) = \max_{b \in B_a}{P_{t}(k,b)}.
  \label{eq:pixel_ceg}
\end{aligned}
\end{equation}

\par By now, we get a new probability map $P^{\prime}_t \in \mathbb{R}^{2 \times H \times W}$. Then, it is easy to obtain the single-temporal segmentation mask $I_t$ and the reliability mask $I^{\textit{rel}}$, which indicates the reliable region. Finally, the pixel-level change mask $I^\textit{pixel-diff}$ is obtained by computing the $l_1$ distance of the bi-temporal segmentation masks. This process can be formulated as:

\begin{equation}
\begin{aligned}
  I_{t}(k) = \textit{argmax}(P^{\prime}_{t}(k)),
  \label{eq:pixel_ceg_it}
\end{aligned}
\end{equation}

\begin{equation}
\begin{aligned}
  I^{\textit{rel}}(k) = \prod \limits_{t=1}^2 \mathds{1}(\max (P^{\prime}_{t}(k)) \geq \gamma),
  \label{eq:pixel_ceg_ig}
\end{aligned}
\end{equation}

\begin{equation}
\begin{aligned}
  I^\textit{pixel-diff}(k) = \lvert I_{1}(k) - I_{2}(k) \rvert,
  \label{eq:pixel_ceg_pd}
\end{aligned}
\end{equation}
where $\textit{argmax}()$ denotes the index of the maximum value, $\mathds{1}$ denotes the indicator function, and $\gamma$ denotes the reliability threshold. $I_{t}$, $I^{\textit{rel}}$ and $I^\textit{pixel-diff}$ are all binary masks whose 0/1 values indicate background/foreground, unreliable/reliable pixels, and unchanged/changed pixels.

\subsubsection{Instance-level CEG}

\begin{figure}[t]
  \centering
   \includegraphics[width=1.0\linewidth]{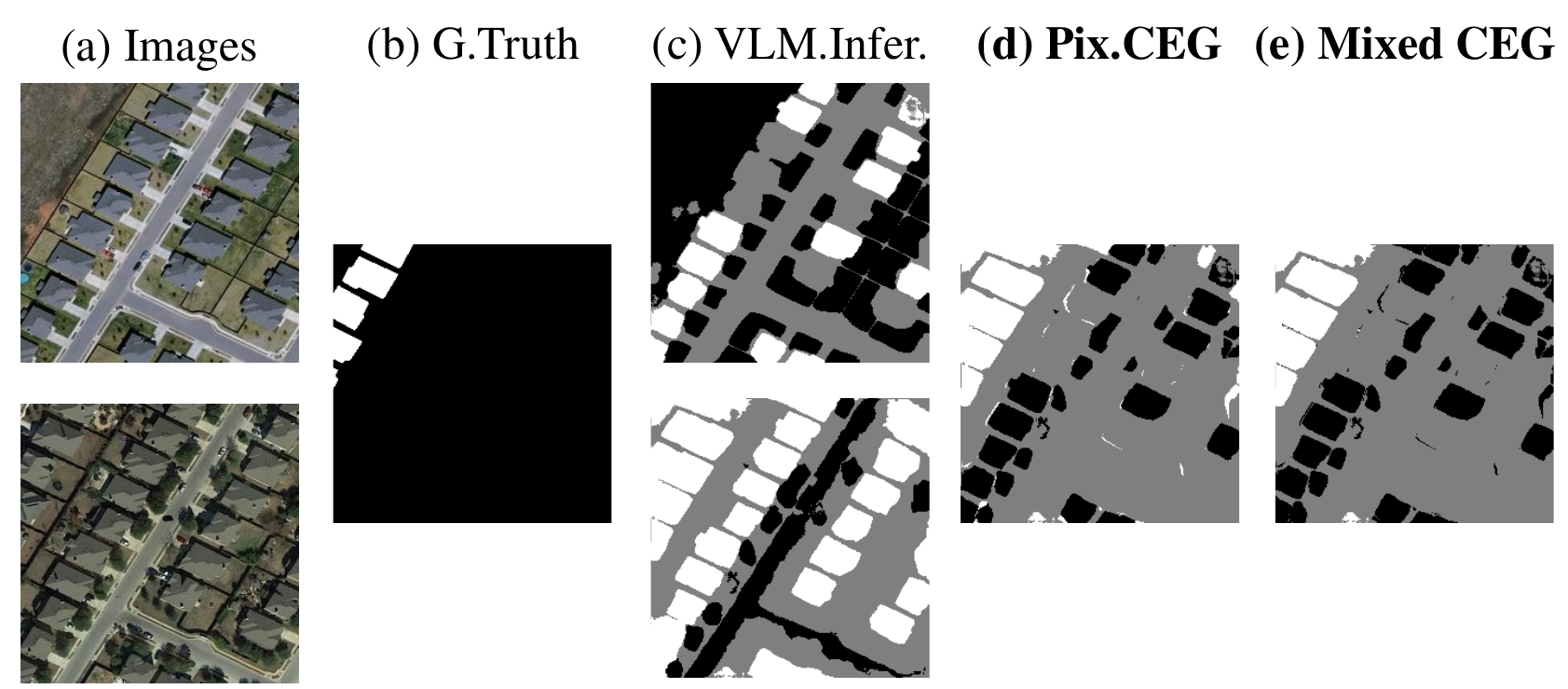}
   \caption{Visualization of the change mask generated by pixel-level CEG and instance-level CEG. The white noise in (d) indicates the non-semantic changes due to object misalignment, which are erased by instance-level CEG in (e). White rendering indicates pixels with semantic changes, black indicates no semantic changes, and gray indicates unreliable regions, which are ignored in the loss computation.}
   \label{fig:fig4}
   \vspace{-1em}
\end{figure}

\par Although pixel-level CEG can delineate high-confidence change regions, however, a significant issue is that in CD, objects in bi-temporal images are not absolutely aligned, e.g., the edges of the object.  This leads to the fact that when using pixel-level CEG, the unaligned regions in the same object of bi-temporal images may present changes in the generated change masks, as shown in Fig. \ref{fig:fig4}(d), which is inconsistent with the concept that CD focuses only on semantic changes. An intuitive solution to this issue is to instantiate the foreground and make instance-level comparisons, i.e., instance-level CEG. This derives two critical questions: how to get instance-level foregrounds and how to make instance-level comparisons. 

\par For the first question, we observe that current segmentation VLMs can be divided into semantic-level models (e.g. MaskCLIP, ZegCLIP, etc.) and instance-level models (e.g. APE, OMG-Seg, etc.). For the former, we can instantiate the connected components for the foreground \cite{suzuki1985topological}; and for the latter, we can directly get the bi-temporal instance mask sets $F_1=\{\textit{ins}^{t_1}_m\}_{m=1}^{M}$ and $F_2=\{\textit{ins}^{t_2}_n\}_{n=1}^{N}$, where $M$, $N$ denote the number of instances in the bi-temporal masks. For the second question, we propose an IoU-aware method to make instance-level comparison in bi-temporal segmentation masks, specifically, we construct a metric matrix $\mathbf{T}$:

\begin{equation}
\begin{aligned}
\mathbf{T}=
\left[\begin{array}{cccc}
\mathbf{s}(\textit{ins}^{t_1}_1, \textit{ins}^{t_2}_1) & \mathbf{s}(\textit{ins}^{t_1}_1, \textit{ins}^{t_2}_2) & \cdots & \mathbf{s}(\textit{ins}^{t_1}_1, \textit{ins}^{t_2}_N) \\
\mathbf{s}(\textit{ins}^{t_1}_2, \textit{ins}^{t_2}_1) & \mathbf{s}(\textit{ins}^{t_1}_2, \textit{ins}^{t_2}_2) & \cdots & \mathbf{s}(\textit{ins}^{t_1}_2, \textit{ins}^{t_2}_N) \\
\vdots & \vdots & \ddots & \vdots \\
\mathbf{s}(\textit{ins}^{t_1}_M, \textit{ins}^{t_2}_1) & \mathbf{s}(\textit{ins}^{t_1}_M, \textit{ins}^{t_2}_2) & \cdots & \mathbf{s}(\textit{ins}^{t_1}_M, \textit{ins}^{t_2}_N), \\
\end{array}\right]
  \label{eq:mm}
\end{aligned}
\end{equation}
where $\mathbf{s}()$ denotes a metric function to describe the similarity of bi-temporal instances, here we use the intersection over union (IoU) as function $\mathbf{s}()$, i.e., 

\begin{equation}
\begin{aligned}
\mathbf{s}(\textit{ins}^{t_1}_m, \textit{ins}^{t_2}_n) = \frac{\textit{ins}^{t_1}_m \cap \textit{ins}^{t_2}_n}{\textit{ins}^{t_1}_m \cup \textit{ins}^{t_2}_n}.
  \label{eq:iou}
\end{aligned}
\end{equation}

\par Then, we take the summation of the metric $\mathbf{s}$ of each instance with all instances on another temporal image as a representation for the probability that the instance is a change event, and from this, we obtain a set $\mathbf{S}$,

\begin{equation}
\begin{aligned}
\mathbf{S} = \{\sum_{m=1}^M\mathbf{s}(\textit{ins}^{t_1}_m, \textit{ins}^{t_2}_1), \cdots,  \sum_{m=1}^M\mathbf{s}(\textit{ins}^{t_1}_m, \textit{ins}^{t_2}_N)\} \cup \\
\{\sum_{n=1}^N\mathbf{s}(\textit{ins}^{t_1}_1, \textit{ins}^{t_2}_n), \cdots,  \sum_{n=1}^N\mathbf{s}(\textit{ins}^{t_1}_M, \textit{ins}^{t_2}_N)\}.
  \label{eq:C}
\end{aligned}
\end{equation}

\par Finally, instances with values in $\mathbf{S}$ smaller than the threshold $\delta$ are assigned as instance-level change events, and their respective binary masks are operated logically ``OR'' with an all-zero mask to obtain the instance-level change mask $I^\textit{ins-diff}$. 


\subsubsection{Mixed CEG}

\par In semi-supervised CD, we expect the pseudo label to be both reliable and reject pseudo changes. From this, mixed CEG is proposed. Specifically, the mixed CEG considers three factors: 1) there is a clear definition of the background in the pixel-level CEG, 2) there are pseudo changes in the pixel-level CEG due to misalignment, and 3) the instance-level CEG focuses only on foreground instances. Thus, the change mask $I^\textit{mix-diff}$ generated by the mixed CEG in the semi-supervised mode is defined as:


\begin{equation}
\begin{aligned}
I^\textit{mix-diff}(k)=\left\{\begin{aligned}
I^\textit{pixel-diff}(k) \cdot  I^\textit{ins-diff}(k), \text{ if } I^\textit{rel}(k)=0 \\
255, \text{ if } I^{rel}(k)=1
\end{aligned}\right.
  \label{eq:I_mix}
\end{aligned}
\end{equation}
where $I^\textit{mix-diff}$ has three value types, i.e., 0 for unchanged regions, 1 for changed regions, and 255 for ignored regions. In this setup, the background definition in pixel-level CEG is effectively exploited and pseudo changes induced by misalignment are eliminated by instance-level CEG. Only the changes recognized by both CEG strategies are retained, ensuring the reliability of the supervised signals.

\subsection{VLM Guidance}
\label{sec:vlm_guidance}

\par The pseudo label $I^\textit{mix-diff}$ generated by VLM can provide additional beneficial guidance for unlabeled data. Under the FixMatch paradigm, for unlabeled data, there are two types of predictions, $y^w$ generated by $x^u$ via weak perturbations and $y^s$ generated via strong perturbations. To fully utilize the pseudo label $I^\textit{mix-diff}$ generated by the VLM, we build supervision from $I^\textit{mix-diff}$ for both $y^w$ and $y^s$ via a CE loss $\mathcal{L}_{vl}$:

\begin{equation}
\begin{aligned}
  \mathcal{L}_{vl}=\frac{1}{B_{u}} \sum \mathcal{H}(I^\textit{mix-diff}, y^s) + \mathcal{H}(I^\textit{mix-diff}, y^w).
  \label{eq:loss_vl}
\end{aligned}
\end{equation}

We argue that regularizing the weak/strong perturbation predictions with a shared VLM label can also be seen as enforcing the consistency between these two predictions. On the other hand, our $\mathcal{L}_{vl}$ shares the core spirits of contrastive learning \cite{yang2023revisiting, he2020momentum, chen2020simple}. Suppose $(q_w, q_s)$ are feature vectors of the weakly/strongly perturbed views of sample $x^u$, and $h_{+}$ is the classifier weight of the class matched to $I^\textit{mix-diff}$. When adopting CE loss, $q_j \cdot h_+$ is maximized against $\sum_{a\in A}{q_j \cdot h_a}$, where $j \in \{w, s\}$, and $h_a$ is classifier weight of concept class $a$. This process is also maximizing the similarity between $q_w$ and $q_s$, which can be approximated as InfoNCE loss \cite{oord2018representation}:

\begin{equation}
\begin{aligned}
\mathcal{L}_{w \leftrightarrow s}=-\log \frac{\exp \left(q_{w} \cdot q_{s}\right)}{\sum_{a\in A} \exp \left(q_{j} \cdot h_{a}\right)} \text {, s.t., } j \in\left\{w, s\right\},
  \label{eq:infonce}
\end{aligned}
\end{equation}
where $q_w$ and $q_s$ are positive pairs, while all other classifier weights except $h_+$ are negative samples\footnote{Note that this formulation is not strictly derived, but simply serves as an intuitive representation of implicitly aligning the two predictions, which is consistent with the insight of contrastive learning. $h_a$ can be considered as the clustering center of all features for class $a$.}.

\subsection{Dual Projection Head}

\par A notable issue is the existence of two types of supervised signals for the prediction $y^s$ of strongly perturbed images, i.e., pseudo labels generated by weakly perturbed images and generated by VLMs. Therefore, a natural problem is that supervised signals from these two different sources may conflict in some regions. We design an extremely simple dual projection head that effectively alleviates this issue at a very slight training cost and obtains a considerable improvement, as shown in Fig. \ref{fig:SemiCD-VL_detail} (3). Specifically, for the output feature $q_s$ of the difference decoder, there exists a linear classifier $\mathbf{h}_{cr}$ for consistency regularization and another linear classifier $\mathbf{h}_{vl}$ for the guidance of VLM:

\begin{equation}
\begin{aligned}
y^s_{cr}=q_{s} \cdot \mathbf{h}_{cr} \\
y^s_{vl}=q_{s} \cdot \mathbf{h}_{vl}
  \label{eq:dph}
\end{aligned}
\end{equation}

\par Then, the hard label $\hat{\mathbf{y}}^{w}$ predicted by the weakly perturbed image is applied to supervise $y^s_{cr}$, as formulated in Eq. \ref{eq:loss_u}; similarly, the change mask $I^\textit{mix-diff}$ generated by the mixed CEG is used to supervise $y^s_{vl}$, in the format of CE loss. Furthermore, since $\mathbf{h}_{vl}$ is independent of $\mathbf{h}_{cr}$, we can also construct supervision from $I^\textit{mix-diff}$ for the prediction $y^w_{vl}$ of weakly perturbed image obtained via $\mathbf{h}_{vl}$, as mentioned in Section \ref{sec:vlm_guidance}.

\subsection{Decoupled Semantic Guidance}

\par In the end-to-end framework, the model directly outputs the change mask of the bi-temporal images, while this process is conducted in a black box. We believe that decoupling the process of change prediction can facilitate the model to understand the CD task more easily, and provides better interpretability. However, in general, decoupling this process requires some auxiliary information, for instance, the respective semantic masks of the bi-temporal images, which requires extra annotation cost. In our framework, the introduction of VLM provides a new possibility for decoupling the CD task, where the respective semantic segmentation masks of the bi-temporal images are also produced during the generation of change masks, as shown in Fig. \ref{fig:SemiCD-VL_detail}. Moreover, to ensure the precision of the segmentation masks, we remove the unreliable prediction regions by a pixel-level threshold $\beta$.

\par In detail, for decoupling change information, two auxiliary segmentation decoders are constructed to respectively predict the bi-temporal segmentation masks, as shown in Fig. \ref{fig:SemiCD-VL_detail} (4). Specifically, these two decoders receive the embeddings of the bi-temporal images from the encoders and up-sample them hierarchically to the high-resolution features $q_{t_1}$ and $q_{t_2}$. Notably, the two segmentation decoders share the same set of weights, making efficient parameter utilization. Finally, the high-resolution bi-temporal features are fed into a linear classifier $\mathbf{h}_{seg}$ to obtain the segmentation prediction $y_{t_1}$ and $y_{t_2}$. These two predictions are supervised by the segmentation masks $I_1$ and $I_2$ (filtered by $\beta$) provided by the VLM via a CE loss and added as part of $\mathcal{L}_{vl}$. The segmentation decoders are activated only during training, without introducing extra cost in reasoning. In the training phase, the bi-temporal segmentation decoders enhance and clarifies the feature representation of the encoder through back-propagation, which explicitly employs single-temporal scene interpretation as a prior task of CD and decouples the bi-temporal entangled features.

\subsection{Contrastive Consistency Regularization}
\label{sec:ccr}

\par In addition to directly using a linear classifier to generate change masks as in Eq. \ref{eq:dph}, another feasible path to CD is the metric learning-based method. Since the basic requirement of CD is pixel-level comparison, it is naturally suited to constrain in the form of contrast. Metric-based CD clusters all positions into two classes by pulling pixel embeddings with the same semantics closer and pushing pixel embeddings with different semantics features, which makes the model capture change representations more efficiently. Specifically, we constrain the high-resolution features $q_{t_1}$ and $q_{t_2}$ from the two segmentation decoders with a contrastive consistency regularization (CCR) loss, and considering the impact of class imbalance, a batch-balanced contrastive loss is applied \cite{chen2020spatial}:

\begin{equation}
\begin{aligned}
\mathcal{L}_{ct}=\left\{\begin{aligned}
\frac{1}{n_{u}} \sum \mathcal{D}\left(q_{t_1}, q_{t_2}\right), \hat{y}=0 \\
\frac{1}{n_{c}} \sum \max \left(0, \epsilon-\mathcal{D}\left(q_{t_1}, q_{t_2}\right)\right), \hat{y}=1
\end{aligned}\right.,
  \label{eq:bcl}
\end{aligned}
\end{equation}
where $n_{u}$, $n_{c}$ denote the number of unchanged pixel pairs and changed pixel pairs in a batch. $\mathcal{D}(,)$ denotes the distance function, here we use pair-wise $l_2$ distance metric. $\epsilon$ denotes the margin, and changed pixel pairs with vector distances greater than $\epsilon$ will not contribute to the loss function. $\hat{y}$ denotes the value of the supervised signal, specifically, we build this contrastive loss for labeled samples and strongly perturbed samples: for the former, $\hat{y}$ denotes the ground truth; and for the latter, $\hat{y}$ denotes the pseudo label predicted by the corresponding weakly perturbed samples.

\subsection{Model Details}

\begin{table}
  \caption{Loss functions of SemiCD-VL. The gray rows indicate the original loss under the consistency regularization framework, i.e., FixMatch \cite{sohn2020fixmatch}.}
  \label{table_loss}
  \centering
  \scalebox{0.9}{
  \begin{tabular}{@{}ll|cccc@{}}
    \toprule[1pt]
    & & labeled & weak pert. & strong pert. \\ 
    \midrule
    \rowcolor{gray!20}
    Supervised Loss & & \checkmark & \ding{55} & \ding{55} \\
    \midrule
    \rowcolor{gray!20}
    Consistency loss & Plain & \ding{55} & \ding{55} & \checkmark \\
    & Contrastive & \checkmark & \ding{55} & \checkmark \\
    \midrule
    VLM Guidance Loss & Change & \checkmark & \checkmark & \checkmark \\
    & Segmentaion & \checkmark & \checkmark & \checkmark \\
    \bottomrule[1pt]
  \end{tabular}}
\end{table}

\par To be consistent with other semi-supervised dense prediction tasks, we use ResNet50 as the encoder and following the general CD setup, the siamese encoder is applied. One decoder is designed for change detection (called difference decoder) and two decoders are designed for single-time segmentation (called segmentation decoder). All decoders have the same network structure, i.e., lightweight all-MLP network \cite{xie2021segformer, fang2023changer}. The pair-wise $l_1$ distances of the bi-temporal features from the siamese encoder are computed to feed into the difference decoder, following the UniMatch setting. For the VLM, APE is applied and we find that it exhibits considerable generalization ability in the chunked RS images. The reason for this might be that the APE is trained directly on higher level fine-grained tasks than the CLIP-derived models \cite{zhou2022extract, zhou2023zegclip}.

\par For the loss function of SemiCD-VL, in addition to the original $\mathcal{L}_{cr}$ in FixMatch, we introduce the VLM guidance loss $\mathcal{L}_{vl}$ and the contrastive loss $\mathcal{L}_{ct}$, as listed in Table \ref{table_loss}. Therefore, the overall loss $\mathcal{L}$ of SemiCD-VL is defined as:

\begin{equation}
\begin{aligned}
\mathcal{L} = \mathcal{L}_{cr} + \lambda_{vl} \mathcal{L}_{vl} + \lambda_{ct} \mathcal{L}_{ct},
  \label{eq:oaloss}
\end{aligned}
\end{equation}
where $\lambda_{vl}$ and $\lambda_{ct}$ denote the weights of the loss terms.

\section{Experiment}
\label{sec:exp}

\subsection{Experiment Setup}

\subsubsection{Dataset}

\par We use two public RSCD datasets, LEVIR-CD and WHU-CD, to validate SemiCD-VL and further validate our CEG strategy, which can be regarded as an un-supervised CD method.

\par - \textbf{The LEVIR-CD dataset} consists of 637 pairs of bi-temporal RS images. Sourced from Google Earth, these images are accompanied by over 31,333 annotated instances of changes. Each image pair has dimensions of $1024 \times 1024$ pixels and a spatial resolution of 0.5 m/pixel.

\par - \textbf{The WHU-CD dataset} contains a pair of images taken in the same area from 2012 and 2016, each with a size of $32,507 \times 15,354$ pixels and a spatial resolution of 0.2 m/pixel.

\par For the LEVIR-CD dataset, we divide it into training, validation, and test sets according to the official division, and all images are cropped into non-overlapping patches of size $256 \times 256$; for the WHU-CD dataset, following SemiCD \cite{bandara2022revisiting}, we crop it into $256 \times 256$ patches and divide it into the training (5947), validation (743), and testing (744) sets. In the semi-supervised mode, partially labeled samples and the remaining unlabeled samples in the training set are used for training, and the results on the testing set are reported.

\subsubsection{Implementation Details}

\par We use PyTorch to build the proposed methods. During training, we use the SGD optimizer and the learning rate is set to 0.02. The confidence threshold $\tau$ for consistency regularization is set to 0.95, following \cite{yang2023revisiting}.
The threshold $\gamma$ and $\beta$ for pixel-level CEG and segmentation pseudo labels are both set to 0.8. The threshold $\delta$ for instance-level CEG is set to 0, which means any intersection is considered unchanged. The loss weights $\lambda_{vl}$ and $\lambda_{ct}$ are set to 0.1, and in particular, we use a linear schedule from 0.1 to 0 for $\lambda_{vl}$ as we consider the VLM guidance at the beginning of training is more important \cite{hoyer2023semivl}. For weak perturbation, random resize-crop and flip are applied, and for strong perturbation, color jitter and CutMix \cite{yun2019cutmix} are applied. For the VLM part, to ensure efficient training, we use VLM to predict the corresponding pseudo labels of all samples beforehand and read them directly in the hard disk during training. All experiments are trained on NVIDIA GeForce RTX 4090 for 80 epochs.

\begin{table*}
      \caption{Comparisons of SemiCD-VL with other semi-supervised methods on LEVIR-CD and WHU-CD testing set with $IoU^c$ (\%) $\uparrow$ metric. All methods are trained on the classic settings, i.e., the labeled images are selected from the original training sets, which consist of 7,120 samples and 5,947 samples, respectively.}
  \label{table_main}
  \centering
  \scalebox{0.9}{
  \begin{tabular}{@{}l|c|cccc|cccc@{}}
    \toprule[1pt]
    \multirow{2}{*}{Method} & \multirow{2}{*}{Backbone} & \multicolumn{4}{c|}{LEVIR-CD(\% labeled)} & \multicolumn{4}{c}{WHU-CD(\% labeled)}\\
    & & 5\%(356) & 10\%(712) & 20\%(1,424) & 40\%(2,848) & 5\%(297) & 10\%(594) & 20\%(1,189) & 40\%(2,378)\\
    \midrule
    SemiCDNet \cite{peng2020semicdnet} & UNet++ & 67.6 & 71.5 & 74.3 & 75.5 & 51.7 & 62.0 & 66.7 & 75.9 \\
    SemiCD \cite{bandara2022revisiting} & ResNet50 & 72.5 & 75.5 & 76.2 & 77.2 & 65.8 & 68.1 & 74.8 & 77.2 \\
    BAN \cite{li2024new} & ResNet50 & 71.3 & 75.4 & 80.3 & 81.6 & 59.2 & 64.9 & 81.1 & 85.3 \\
    \midrule
    SemiVL \cite{hoyer2023semivl} & ViT-B/16 & 71.6 & 74.3 & - & - & - & - & - & - \\
    \midrule
    AdvNet \cite{vu2019advent} & ResNet101 & 66.1 & 72.3 & 74.6 & 75.0 & 55.1 & 61.6 & 73.8 & 76.6 \\
    s4GAN \cite{mittal2019semi} & ResNet101 & 64.0 & 67.0 & 73.4 & 75.4 & 18.3 & 62.6 & 70.8 & 76.4 \\
    Sup. only $|$ PSPNet & ResNet50 & 67.5 & 73.4 & 75.2 & 77.7 & 48.3 & 60.7 & 69.7 & 69.5 \\
    UniMatch $|$ PSPNet \cite{yang2023revisiting} & ResNet50 & 75.6 & 79.0 & 79.0 & 78.2 & 77.5 & 78.9 & 82.9 & 84.4 \\
    Sup. only $|$ DeepLabV3 & ResNet50 & 69.3 & 76.0 & 77.6 & 80.5 & 54.1 & 60.9 & 68.4 & 76.2 \\
    UniMatch $|$ DeepLabV3 \cite{yang2023revisiting} & ResNet50 & 80.7 & 82.0 & 81.7 & 82.1 & 80.2 & 81.7 & 81.7 & 85.1 \\
    \midrule
    FixMatch \cite{sohn2020fixmatch} & ResNet50 & 79.5 & 81.3 & 81.9 & 81.9 & 76.5 & 80.6 & 81.0 & 81.8 \\
    SemiCD-VL & ResNet50 & \textbf{81.9} \textbf{{\color{green!50!black}$\uparrow$\scriptsize{2.4}}} & \textbf{82.6} \textbf{{\color{green!50!black}$\uparrow$\scriptsize{1.3}}} & \textbf{82.7} \textbf{{\color{green!50!black}$\uparrow$\scriptsize{0.8}}} & \textbf{83.0} \textbf{{\color{green!50!black}$\uparrow$\scriptsize{1.1}}} & \textbf{81.8} \textbf{{\color{green!50!black}$\uparrow$\scriptsize{5.3}}} & \textbf{83.2} \textbf{{\color{green!50!black}$\uparrow$\scriptsize{2.6}}} & \textbf{84.8} \textbf{{\color{green!50!black}$\uparrow$\scriptsize{3.8}}} & \textbf{85.7} \textbf{{\color{green!50!black}$\uparrow$\scriptsize{3.9}}} \\
    \bottomrule[1pt]
  \end{tabular}}
  \vspace{-1em}
\end{table*}

\subsubsection{Evaluation metrics}

\par We use the $\textit{IoU}^c$ and $F_1^c$ score as evaluation metrics, which are calculated as follows:

\begin{equation}
\label{IoU_c}
\textit{IoU}^c = \frac{TP}{TP+FP+FN},
\end{equation}

\begin{equation}
\label{F1-Score}
F_1^c = \frac{2TP}{2TP+FP+FN},
\end{equation}
where TP, FP and FN indicate true positive, false positive, and false negative, which are calculated on the change category to avoid the class imbalance problem. 

\subsection{Main Results}

\par We select several comparison methods from different perspectives, as listed in Table \ref{table_main}. SemiCDNet \cite{peng2020semicdnet} and SemiCD \cite{bandara2022revisiting} are specialized SSL models for RSCD. SemiCDNet is built under the adversarial SSL framework, which contains two discriminative networks, one to determine whether the segmentation output is either from unlabeled images or from the ground truth, and the other to encourage similarity between the entropy maps of unlabeled and labeled samples. SemiCD is a consistency regularity-based method that builds learning of unlabeled samples by adding random feature perturbations to the encoder's difference features and forcing their predictions to be consistent. Another worthy comparison is BAN \cite{li2024new}, which is a parameter-efficient fine-tuning RSCD framework based on the foundation model, rather than an SSL method, but with excellent transfer capabilities. When the labeled samples are relatively rich (20\%, 40\%), BAN shows superiority over the previous two RSCD semi-supervised methods, but its performance drops drastically when the samples are fewer (5\%, 10\%), which illustrates the necessity of learning on unlabeled samples.

\par SemiVL \cite{hoyer2023semivl} is a SOTA semi-supervised segmentation model based on VLM guidance and can be seen as one of the baselines for SemiCD-VL. We modify SemiVL to fit the CD task, however, it does not show satisfactory performance. We speculate that there are three reasons: 1) the drastic down-sampling in ViT backbones leads to the inability to achieve fine segmentation in RSCD, 2) the lack of a reasonable CEG strategy, and 3) the lack of elaborate design for the CD task.

\par
\begin{figure*}[t]
  \centering
   \includegraphics[width=0.9\linewidth]{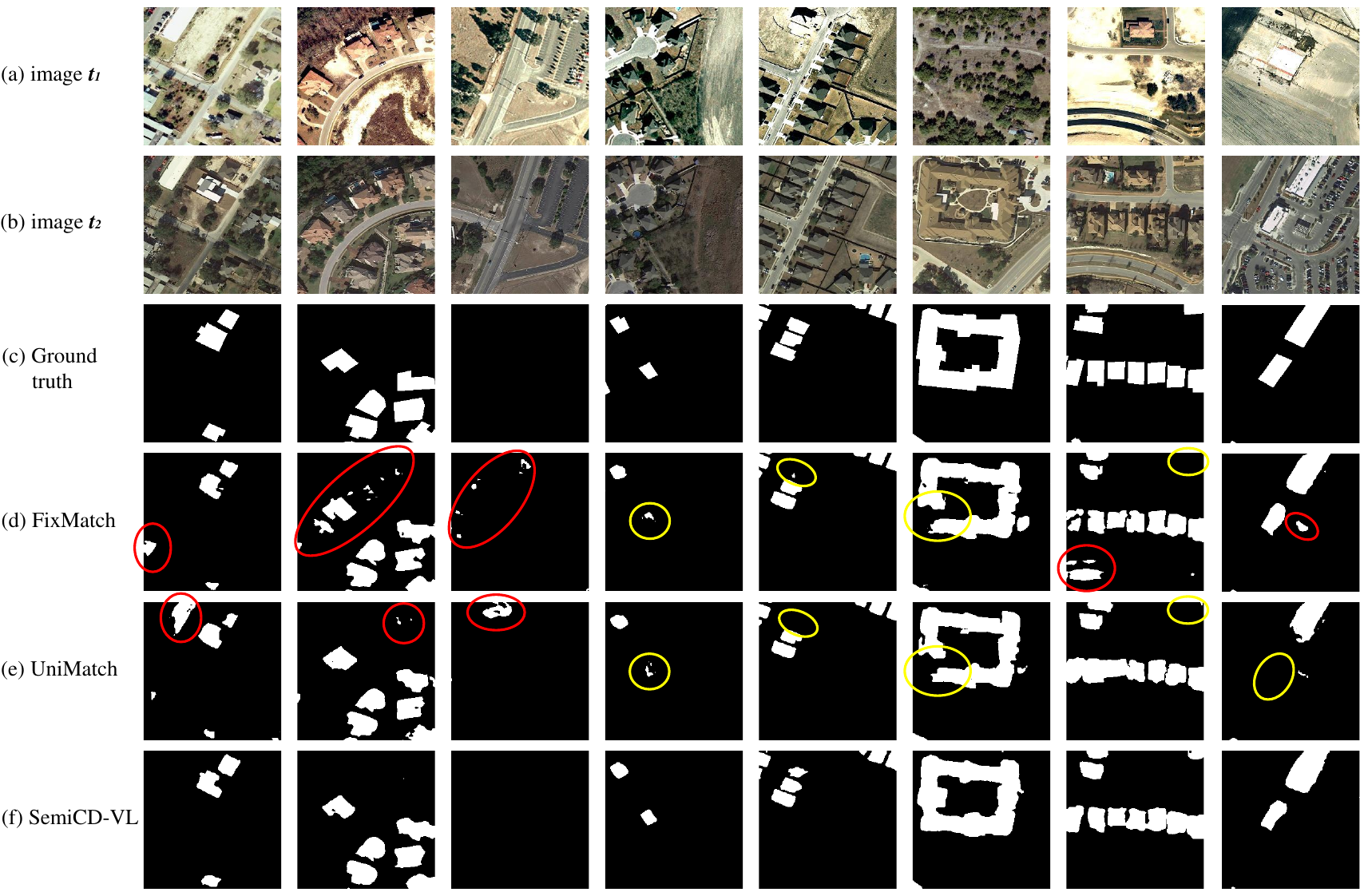}
   \caption{Visualization results of different semi-supervised CD methods on the LEVIR-CD dataset with 5\% labels. Red circle indicates the false alarm and yellow circle indicates the missed detection.}
   \label{fig:vis}
   \vspace{-1em}
\end{figure*}

 In addition, some common semi-supervised semantic segmentation models are modified for CD tasks here, in brief, their backbones are reconstructed to siamese structures. AdvNet \cite{vu2019advent} and s4GAN \cite{mittal2019semi} are both adversarial-based models, the former directly or indirectly makes the information entropy of unlabeled data lower, and the latter expects a more similar distribution between labeled and unlabeled samples. FixMatch \cite{sohn2020fixmatch} combines consistency regularization and pseudo labeling while vastly simplifying the overall approach, which we have described in detail in Section \ref{sec:Preliminaries}. UniMatch \cite{yang2023revisiting} adds additional streams of image-level perturbations and feature-level perturbations to FixMatch and supervises them simultaneously using pseudo labels generated by the weak perturbations. On RSCD datasets, FixMatch and UniMatch show decent performance, superior to adversarial-based models. Our method, starting from FixMatch, obtains significant performance gains and achieves optimal $\textit{IoU}^c$ score. Specifically, on the LEVIR-CD dataset, 81.9\% $\textit{IoU}^c$ is achieved when using only 5\% labeled data. The improvement on the WHU-CD dataset is more obvious, where the $\textit{IoU}^c$ score improves from 76.5\% to 81.8\%, a span of 5.3\%, when using 5\% labeled data for training. And the improvement is persistent as the labeled data volume increases. Notably, the components in SemiCD-VL do not conflict with UniMatch, and their combination can yield further enhancements, but to make SemiCD-VL clearer, we do not elaborate further coupling with other methods. In Fig. \ref{fig:vis}, some visual comparisons of the three methods are presented. It is observed that SemiCD-VL improves in terms of precision and recall, reducing false alarms and missed detections. This implies that the supervision of VLM facilitates the model to distinguish between foreground and background.

\begin{table}
  \caption{Comparisons of SemiCD-VL with supervised CD methods.}
  \label{table_sup}
  \centering
  \scalebox{0.9}{
  \begin{tabular}{@{}l|cc|cc@{}}
    \toprule[1pt]
    \multirow{2}{*}{Method} & \multicolumn{2}{c|}{LEVIR-CD} & \multicolumn{2}{c}{WHU-CD}\\
    & $F_1^c$ & $IoU^c$ & $F_1^c$ & $IoU^c$\\
    \midrule
    100\% labeled data: & & & & \\
    FC-EF \cite{daudt2018fully} & 83.4 & 71.5 & 69.4 & 53.1 \\
    FC-Siam-Diff \cite{daudt2018fully} & 86.3 & 75.9 & 58.8 & 41.7 \\
    FC-Siam-Conc \cite{daudt2018fully} & 83.7 & 72.0 & 66.6 & 50.0 \\
    STANet \cite{chen2020spatial} & 87.3 & 77.4 & 82.3 & 70.0 \\
    SNUNet \cite{fang2021snunet} & 88.2 & 78.8 & 83.5 & 71.7 \\
    BiT \cite{chen2021remote} & 89.3 & 80.7 & 84.0 & 72.4 \\
    ChangeFormer \cite{bandara2022transformer} & 90.4 & 82.5 & 89.9 & 81.6 \\
    \midrule
    SemiCD-VL (5\% labeled data) & 90.1 & 81.9 & 90.0 & 81.8 \\
    SemiCD-VL (10\% labeled data) & \textbf{90.5} & \textbf{82.6} & \textbf{90.8} & \textbf{83.2} \\
    \bottomrule[1pt]
  \end{tabular}}
  \vspace{-1em}
\end{table}

\par In Table \ref{table_sup}, we list some supervised CD models that are trained on the entire training set. It can be observed that when using only 5\% of the labeled data, SemiCD-VL's performance is close to that of the models trained with all labels. While using 10\% of the labeled data, SemiCD-VL outperforms the best supervised CD models. These results demonstrate that our VLM guidance strategy makes the semi-supervised and fully supervised CD methods comparable.

\subsection{Cross-dataset Results}

\begin{table}
  \caption{Comparisons of SemiCD-VL and other methods under cross-dataset condition. All methods are trained on 5\% or 10\% labeled data from the LEVIR-CD dataset and unlabeled data from the WHU-CD dataset, and the reported results are evaluated on the LEVIR-CD testing set.}
  \label{table_cross_dataset}
  \centering
  \scalebox{0.9}{
  \begin{tabular}{@{}l|ccc@{}}
    \toprule[1pt]
    Method & 5\%(356) & 10\%(712)\\
    \midrule
    AdvNet \cite{vu2019advent} & 65.2 & 71.6 \\
    s4GAN \cite{mittal2019semi} & 66.7 & 68.4 \\
    SemiCDNet \cite{peng2020semicdnet} & 67.6 & 71.9 \\
    SemiCD \cite{bandara2022revisiting} & 71.4 & 74.6 \\
    FixMatch \cite{sohn2020fixmatch} & 75.1 & 78.3 \\
    UniMatch \cite{yang2023revisiting} & 77.0 & 78.3 \\
    \midrule
    SemiCD-VL & \textbf{78.1} & \textbf{80.3} \\
    \bottomrule[1pt]
  \end{tabular}}
  \vspace{-1em}
\end{table}

\par To verify whether SemiCD-VL has the ability to improve CD by utilizing unlabeled data from other datasets, we conduct cross-dataset experiments on LEVIR-CD and WHU-CD datasets. Specifically, we select some samples from the LEVIR-CD dataset as labeled data and images from the WHU-CD dataset as unlabeled data, and evaluate models on the testing set of LEVIR-CD, as listed in Table \ref{table_cross_dataset}. Compared to our baseline model FixMatch, SemiCD-VL gets 3.0\% and 2.0\% improvement to achieve 78.\% and 80.3\% $\textit{IoU}^c$ scores when using 5\% and 10\% labeled data. This experiment demonstrates the potential of SemiCD-VL to learn from infinite cross-domain data.

\subsection{Un-supervised Results}
\label{sec:unsup_res}

\par Our CEG strategy can be directly used as an un-supervised CD method. We compare the instance-level CEG with several traditional and SOTA un-supervised CD methods, as listed in Table \ref{table_unsup}. Notably, different from earlier methods, DINOv2+CVA \cite{zheng2024segment}, AnyChange \cite{zheng2024segment} and SCM \cite{tan2023segment} utilize the power of the foundation model. Specifically, DINOv2+CVA extracts the bi-temporal image embeddings using the pre-trained DINOv2 \cite{oquab2023dinov2}, computes the $l_2$ norm of their difference, and generates the binary mask using an adaptive threshold. AnyChange utilizes the capability of the segment anything model (SAM) \cite{kirillov2023segment} to infer instance masks and retrieve their corresponding feature regions, and then computes pixel-level or instance-level feature similarity by cosine distance. However, unfortunately, AnyChange is not semantic-aware since SAM is class-agnostic. Correspondingly, SCM introduces CLIP in the un-supervised CD pipeline to filter out non-specified changes by foreground text guidance. They have achieved a significant improvement over previous methods, benefiting from the accumulated knowledge of the foundation model. Beyond them, our method reaches a new milestone in un-supervised CD, driven by VLM and instance-level CEG. On the testing datasets, our instance-level CEG achieves more than double the improvement over the previous best methods, with $\textit{IoU}^c$ improved from 18.8\% to 46.3\% on the LEVIR-CD dataset and from 18.6\% to 45.2\% on the WHU-CD dataset. Furthermore, functionally, the instance-level CEG not only generates change masks, but also yields single-temporal semantic masks, which is significant for downstream tasks (e.g., semantic change detection).

\begin{table}
  \caption{Comparisons of our instance-level CEG with other un-supervised CD methods on LEVIR-CD and WHU-CD datasets.}
  \label{table_unsup}
  \centering
  \scalebox{0.9}{
  \begin{tabular}{@{}c|cc|cc@{}}
    \toprule[1pt]
    \multirow{2}{*}{Method} & \multicolumn{2}{c|}{LEVIR-CD} & \multicolumn{2}{c}{WHU-CD} \\
    & $IoU^c$ & $F_1^c$ & $IoU^c$ & $F_1^c$ \\ 
    \midrule
    PCA-KM \cite{celik2009unsupervised} & 4.8 & 9.1 & 5.4 & 10.2 \\
    CNN-CD \cite{el2016convolutional} & 7.0 & 13.1 & 4.9 & 9.4 \\
    DSFA \cite{du2019unsupervised} & 4.3 & 8.2 & 4.1 & 7.8 \\
    DCVA \cite{saha2019unsupervised} & 7.6 & 14.1 & 10.9 & 19.6 \\
    GMCD \cite{tang2021unsupervised} & 6.1 & 11.6 & 10.9 & 19.7 \\
    DINOv2+CVA \cite{zheng2024segment} & - & 17.3 & - & - \\
    AnyChange-H \cite{zheng2024segment} & - & 23.0 & - & - \\
    SCM \cite{tan2023segment} & 18.8 & 31.7 & 18.6 & 31.3 \\
    \midrule
    \textbf{Instance-level CEG (ours)} & \textbf{46.3} & \textbf{63.3} & \textbf{45.2} & \textbf{62.3} \\
    \textbf{$\Delta$} & \textbf{{\color{green!50!black}$\uparrow$27.5}} & \textbf{{\color{green!50!black}$\uparrow$31.6}} & \textbf{{\color{green!50!black}$\uparrow$26.6}} & \textbf{{\color{green!50!black}$\uparrow$31.0}} \\
    \bottomrule[1pt]
  \end{tabular}}
  \vspace{-1em}
\end{table}

\begin{table}
  \caption{Comparison of pixel-level CEG and instance-level CEG on LEVIR-CD dataset. ``Total'' denotes the evaluation metrics over all pixels, ``Valid'' denotes the evaluation metrics over reliable pixels only, i.e., when the probability is greater than the threshold $\gamma$, and ``ratio'' denotes the ratio of reliable pixels.}
  \label{table_ablation_unsup}
  \centering
  \scalebox{0.9}{
  \begin{tabular}{@{}cc|cc|ccc@{}}
    \toprule[1pt]
    \multirow{2}{*}{Pixel-level ($\gamma$)} & \multirow{2}{*}{Instance-level} & \multicolumn{2}{c|}{Total} & \multicolumn{3}{c}{Valid} \\
    & & $IoU^c$ & $F_1^c$ & $IoU^c$ & $F_1^c$ & ratio \\ 
    \midrule
    0.1 & w/o & 36.9 & 53.9 & 41.2 & 58.4 & 71.9 \\
    0.5 & w/o & 41.4 & 58.5 & 54.4 & 70.5 & 55.4 \\
    0.8 & w/o & 31.7 & 48.2 & 65.7 & 79.3 & 34.4 \\
    w/o & w & \textbf{46.3} & \textbf{63.3} & - & - & - \\
    0.8 & w & 32.3 & 48.8 & \textbf{69.3} & \textbf{81.9} & 34.4 \\
    \bottomrule[1pt]
  \end{tabular}}
  \vspace{-1em}
\end{table}

\par In Table \ref{table_ablation_unsup}, we compare pixel-level CEG and instance-level CEG. When using pixel-level CEG, the best total performance is achieved with the threshold $\gamma$ set to 0.5, obtaining 41.4\% and 58.5\% in $\textit{IoU}^c$ and $F_1^c$ scores. And when using instance-level CEG (threshold $\delta$ defaulted to 0), the $\textit{IoU}^c$ and $F_1^c$ scores are improved to 46.3\% and 63.3\%, which demonstrates the significant advantage of the instance-level CEG under the purely un-supervised mode.

\subsection{Ablation Studies}

\subsubsection{Effectiveness of Mixed CEG}

\begin{table}
  \caption{Ablation study of SemiCD-VL’s components on LEVIR-CD (5\% labeled data): vision-language model guidance (VLM.Guid.), Mixed CEG, dual projection head (DP.Head), decoupled semantic guidance (Dec.Guid.), and contrastive consistency regularization (CCR).}
  \label{table_ablation}
  \centering
  \scalebox{0.9}{
  \begin{tabular}{@{}ccccc|cc@{}}
    \toprule[1pt]
    VLM.Guid. & Mixed CEG & DP.Head & Dec.Guid. & CCR & \multicolumn{2}{c}{$IoU^c$} \\ 
    \midrule
    - & - & - & - & - & 79.52 & $\Delta$ \\
    \checkmark & - & - & - & - & 80.66 & {\color{green!50!black}$\uparrow$1.14} \\
    \checkmark & \checkmark & - & - & - & 80.97 & {\color{green!50!black}$\uparrow$1.45} \\
    \checkmark & \checkmark & \checkmark & - & - & 81.23 & {\color{green!50!black}$\uparrow$1.71} \\
    \checkmark & \checkmark & \checkmark & - & \checkmark & 81.15 & {\color{green!50!black}$\uparrow$1.63} \\
    \checkmark & \checkmark & \checkmark & \checkmark & - & 81.65 & {\color{green!50!black}$\uparrow$2.13} \\
    \checkmark & \checkmark & \checkmark & \checkmark & \checkmark & 81.94 & {\color{green!50!black}$\uparrow$2.42} \\
    \bottomrule[1pt]
  \end{tabular}}
\end{table}

\par As mentioned in Section \ref{sec:mix_ceg}, unlike in the pure un-supervised mode, in the semi-supervised mode we focus more on the precision of the pixels selected to validly provide supervised signals for the unlabeled data, and thus introduce the mixed CEG. In Table \ref{table_ablation_unsup}, we directly evaluate the reliable (valid) pixel portion on the testing set ("Valid" columns). As the pixel-level threshold $\gamma$ is higher, the precision in the reliable pixels improves, which indicates that the pixel threshold $\gamma$ is effective in filtering out the low-precision pixels. However, naturally, the total number of pixels retained decreases as the threshold is increased, implying sparser supervised signals. When both pixel-level CEG and instance-level CEG, i.e., mixed CEG, are used, the highest precision of reliable pixels is achieved, with an improvement in $\textit{IoU}^c$ from 65.7\% to 69.3\%, and the ratio of effective pixels does not decrease, compared to using only pixel-level CEG with a threshold of 0.8.

\subsubsection{Components of SemiCD-VL}

\begin{table}
  \caption{Ablation study of thresholds (pixel-level) for the generation of pseudo change labels and segmentation labels.}
  \label{table_ablation_thres}
  \centering
  \scalebox{0.9}{
  \begin{tabular}{@{}cc|cccc@{}}
    \toprule[1pt]
    $\gamma$ & $\beta$ & Presicion & Recall & $IoU^c$ \\
    \midrule
    0.9 & 0.9 & 92.9 & 87.3 & 81.8 \\
    0.8 & 0.8 & 92.8 & 87.5 & \textbf{81.9} \\
    0.5 & 0.5 & 92.2 & 87.9 & 81.8 \\
    0.1 & 0.1 & 92.7 & 87.1 & 81.5 \\
    0.8 & 0.5 & 93.9 & 85.7 & 81.2 \\
    0.5 & 0.8 & 92.9 & 86.6 & 81.2 \\
    0.8 & 0.1 & 93.3 & 86.3 & 81.3 \\
    0.1 & 0.8 & 92.3 & 97.1 & 81.2 \\
    \bottomrule[1pt]
  \end{tabular}}
  \vspace{-1em}
\end{table}

\begin{table}
  \caption{Ablation study of weights $\lambda_{ct}$ and $\lambda_{vl}$ for contrastive loss and VLMguidance loss.}
  \label{table_ablation_weight}
  \centering
  \scalebox{0.9}{
  \begin{tabular}{@{}cc|cccc@{}}
    \toprule[1pt]
     $\lambda_{ct}$ & $\lambda_{vl}$ & Presicion & Recall & $IoU^c$ \\
    \midrule
    0.1 & 1.0 & 93.1 & 86.6 & 81.4 \\
    0.1 & 0.5 & 92.8 & 87.1 & 81.6 \\
    0.1 & 0.1 & 92.8 & 87.5 & \textbf{81.9} \\
    0.1 & 0.01 & 94.0 & 85.8 & 81.3 \\
    1.0 & 0.1 & 91.8 & 71.0 & 66.8 \\
    0.5 & 0.1 & 93.9 & 84.0 & 79.7\\
    0.01 & 0.1 & 92.6 & 87.5 & 81.8 \\
    \bottomrule[1pt]
  \end{tabular}}
  \vspace{-1em}
\end{table}

\par Compared to the FixMatch baseline, five components are introduced in SemiCD-VL, and in Table \ref{table_ablation}, we analyze them in detail. Additional supervised signals are introduced through the VLM guidance, and the $\textit{IoU}^c$ is improved by 1.14\% compared to FixMatch. After replacing the pixel-level CEG with the mixed CEG, the performance is improved to 80.97\%, which indicates that removing the pseudo label noise (i.e., unaligned pseudo-change regions) is effective. The dual projection head avoids the conflict of supervised signals from different sources and further improves the model performance. Notably, the performance decreases after applying CCR, which we attribute to the insufficient guidance to the segmentation decoder. After applying the VLM guidance to the segmentation decoder, the $\textit{IoU}^c$ score increases from 81.15\% to 81.94\%, which again confirms the significance of the VLM guidance. Although five components are added to FixMatch, all of them are activated only during training and do not bring extra reasoning costs.

\subsubsection{Hyper-parameter Adjustment}

\par In Table \ref{table_ablation_unsup}, we observe that the performance in un-supervised mode has significant ebb and flow as the pixel-level threshold is changed, so it is essential to explore the effect of these thresholds on SemiCD-VL in semi-supervised mode. As listed in Table \ref{table_ablation_thres}, we find that SemiCD-VL is insensitive to the threshold of the pseudo labels generated by VLM, with only 0.1\% undulation in $\textit{IoU}^c$ over the threshold change range from 0.5 to 0.9. In addition, an interesting observation is that the model prefers consistent CD threshold $\gamma$ and segmentation threshold $\beta$, partially due to CCR, which encourages consistency in segmentation and CD predictions. In fact, this is a side evidence that the CD process is decoupled and that single-temporal scene interpretation and bi-temporal change detection are endowed with causality.

\par In Table \ref{table_ablation_weight}, we conduct ablation studies for the VLM guidance loss $\mathcal{L}_{vl}$ and the contrastive loss $\mathcal{L}_{ct}$. It can be observed that both types of losses can bring a certain degree of gain to SemiCD-VL, however, neither of them can be the major contribution term to the overall loss. Especially for the contrastive loss, when the weight of $\mathcal{L}_{ct}$ is set to greater than 0.5, the model performance decreases dramatically. For the VLM guidance loss, this phenomenon exists as well, but is relatively slight. We believe that this is caused by the inevitable presence of errors in the pseudo labels generated by the VLM, which are only capable of performing the auxiliary terms of the CD under the current capabilities of VLM. Further, this explains why we introduce the VLM into the SSL framework instead of direct un-supervised reasoning.

\section{Conclusion}
\label{sec:concl}

\par In this paper, we propose a VLM guidance-based semi-supervised CD method SemiCD-VL, aiming to explore the effectiveness of VLM in CD tasks. To sufficiently and rationally utilize the pseudo labels generated by VLM, we introduce five components in SemiCD-VL, which not only provide abundant and reliable pseudo labels for unlabeled data, but also explicitly decouple the CD process. We conduct extensive experiments on two RSCD datasets and the results demonstrate the superiority of SemiCD-VL. In addition, our proposed CEG strategy makes a performance leap in un-supervised CD. Further, SemiCD-VL is also applicable to a wide range of CD tasks in other scenarios.

\par There are two main limitations of SemiCD-VL: 1) Although we propose the mixed CEG strategy to obtain more reliable pseudo labels, it is inevitable that the incorrect pseudo labels still exist due to the imperfect VLMs. In addition, the current pseudo-label generation pipeline, i.e., ``single-temporal reasoning + post-processing'', suffers from an inherent flaw, i.e., the accumulation of errors due to multi-step processing. This issue is expected to be addressed by future end-to-end multi-temporal VLMs. 2) The generation of pseudo labels is time-consuming, which requires extra training cost, or pre-generating the required VLM pseudo labels. The merit of this research is that it demonstrates the possibilities and potential of VLMs in semi-supervised and un-supervised CD tasks, which is a direct or indirect path to realize a universal CD model in the future.



\section*{Acknowledgements}

The authors would like to thank Xiaoliang Tan and Guanzhou Chen for some experimental data on un-supervised CD methods (Table \ref{table_unsup}).

\bibliographystyle{IEEEtran}
\bibliography{IEEEabrv,references.bib}

\end{document}